\documentclass[10pt,twocolumn,letterpaper]{article}

\usepackage{cvpr}
\usepackage{times}
\usepackage{epsfig}
\usepackage{graphicx}
\usepackage{amsmath}
\usepackage{amssymb}
\usepackage{booktabs}
\usepackage{multirow}
\usepackage{multicol}
\usepackage{pifont}

\usepackage[pagebackref=true,breaklinks=true,letterpaper=true,colorlinks,bookmarks=false]{hyperref}

\cvprfinalcopy % *** Uncomment this line for the final submission

\def\cvprPaperID{**} % *** Enter the CVPR Paper ID here

\ifcvprfinal\fi
\begin{document}

%%%%%%%%% TITLE
\title{Self-supervised Object Motion and Depth Estimation from Video}

\author{
Qi Dai\textsuperscript{1,3}\space\space\space\space Vaishakh Patil\textsuperscript{1}\space\space\space\space Simon Hecker\textsuperscript{1}\space\space\space\space Dengxin Dai\textsuperscript{1}\space\space\space\space Luc Van Gool\textsuperscript{1,2}\space\space\space\space Konrad Schindler\textsuperscript{1,3} \vspace{6px} \\
\textsuperscript{1}Computer Vision Lab, ETH Zurich\space\space\space\space \textsuperscript{2}VISICS, ESAT/PSI, KU Leuven\\
\textsuperscript{3}Institute of Geodesy and Photogrammetry, ETH Zurich \\
{\tt\small daiq@ethz.ch $\{$patil, heckers, dai, vangool$\}$@vision.ee.ethz.ch schindler@geod.baug.ethz.ch}
}

\maketitle
%\thispagestyle{empty}

%%%%%%%%% ABSTRACT
\begin{abstract}

We present a self-supervised learning framework to estimate the individual object motion and monocular depth from video. We model the object motion as a 6 degree-of-freedom rigid-body transformation. The instance segmentation mask is leveraged to introduce the information of object. Compared with methods which predict dense optical flow map to model the motion, our approach significantly reduces the number of values to be estimated. Our system eliminates the scale ambiguity of motion prediction through imposing a novel geometric constraint loss term. Experiments on KITTI driving dataset demonstrate our system is capable to capture the object motion without external annotation. Our system outperforms previous self-supervised approaches in terms of 3D scene flow prediction, and contribute to the disparity prediction in dynamic area.

\end{abstract}

%%%%%%%%% BODY TEXT
\section{INTRODUCTION}

Imagining a driving scenario in real world. The driver may encounter many dynamic objects (\textit{e.g.} moving vehicles). The knowledge of their movements is of vital importance for the driving safety. We aim to solve the motion of individual object from video in the context of autonomous driving (\textit{i.e.} the video is taken by a camera installed on a moving car). However, due to the entanglement of object movement and camera ego-motion, it is difficult to estimate the individual object motion from video. 

This difficulty can be tackled by introducing the information of surrounding structure, \textit{i.e.} a per-pixel depth map. Depth estimation from image is a fundamental problem in computer vision. Recently the view-synthesis based approach provides a self-supervised learning framework for depth estimation, without supervision of depth annotation. Strong baselines of depth prediction have been established in \cite{godard2017unsupervised, luo2018every, zhou2017unsupervised}, most of which jointly train a depth and a pose network (for predicting camera ego-motion). 

\begin{figure}[!t]
\begin{center}
    \includegraphics[width=1.0\linewidth]{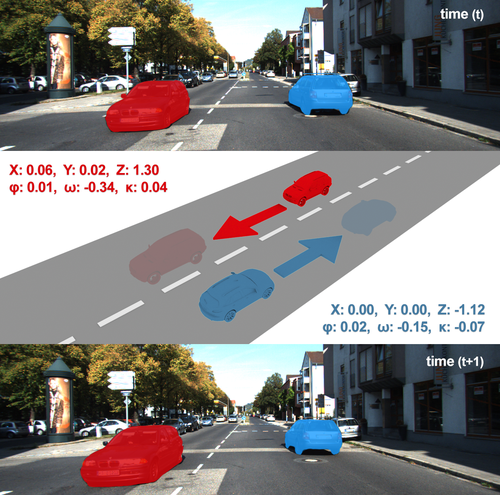}
\end{center}
    \caption{Our system predicts individual object motion by leveraging the instance-level segmentation mask. For each segmented object, three translation (X, Y, Z) and three rotation elements ($\varphi$, $\omega$, $\kappa$) are predicted. The prediction describes the object movement during the capture of two consecutive frames ($I_t$ and $I_{t+1}$), within the camera coordinate system of $I_t$. The unit of translation and rotation elements are meter and degree respectively. }
  \vspace{-4mm}
\label{fig:intro}
\end{figure}

The depth and camera ego-motion can only explain the pixel displacement in static background. To explain the motion of dynamic object, 2D optical flow (\cite{yin2018geonet}) and 3D scene flow map (\cite{sfCaoKHM2019}) have been used to model the object motion. For example, Luo \textit{et al.} \cite{luo2018every} proposed to jointly estimate depth, camera ego-motion and optical flow map. 

In this paper, we propose a self-supervised learning framework for estimating the individual object motion and the monocular depth from video. The object motion is modelled in the form of a 6 degree-of-freedom (\textit{dof}) rigid-body transform. We further eliminate the scale ambiguity of motion prediction by imposing a novel geometric constraint loss term. Previous approaches use dense flow map to model the motion, meaning a pixel-wise flow map is predicted. By contrast, our approach predicts a 6 \textit{dof} rota-translation for the motion of individual object. The number of values to be estimated is significantly reduced from a pixel-wise prediction to 6 scalars per instance. 

We perform evaluations of our framework on KITTI dataset. The result manifests the effectiveness of our system to predict individual object motion. Our system outperforms other self-supervised approaches in scene flow prediction, and improve the disparity prediction in dynamic area of the image. 

\section{Related Work}

Our system is developed to solve the individual object motion from video, and provide monocular depth estimation. In this section we firstly present works related to depth estimation from image. Then some methods which address the object motion are introduced. \\

\noindent {\bf Supervised Depth Estimation}\, The depth estimation is formulated as a regression problem in most supervised approaches, where the difference between the predicted depth and its ground truth is minimized. The manually defined feature is used in early work. Saxena \textit{et al.} \cite{saxena2006learning} propose to estimate the single-view depth by training Markov random field(MRF) with hand-crafted features. Liu \textit{et al.} \cite{liu2010single} integrate semantic labels with MRF learning. Ladicky \textit{et al.} \cite{ladicky2014pulling} improve the performance by combining the semantic labeling with the depth estimation. 

Deep convolutional neural network (CNN) is good at extracting features and inspires many other methods. Eigen \textit{et al.} \cite{eigen2014depth} propose a CNN architecture to produce dense depth map. Based on this architecture, many variants have been proposed to improve the performance. Li \textit{et al.} \cite{li2015depth} improve the estimation accuracy by combining the CNNs with the conditional random filed(CRF), while Laina \textit{et al.} \cite{laina2016deeper} use the more robust \textit{Huber loss} as the loss function. Patil~\etal~\cite{patil2020don} produce a more accurate depth estimation by exploiting spatio-temporal structures of depth across frames. 

\noindent {\bf Self-supervised Depth Estimation}\, The depth map can be learned from unlabeled video under a view-synthesis based framework \cite{zhou2017unsupervised}. This framework is primarily supervised by the image reconstruction loss, which is a function of depth prediction. Zhou \textit{et al.} \cite{zhou2017unsupervised} proposed to jointly train two networks for estimating dense depth and camera ego-motion, respectively. The image is synthesized from the network outputs, following the traditional \textit{Structure-from-motion} procedure. Extra constraint and additional information have been introduced to improve the performance, like the temporal depth consistency \cite{mahjourian2018unsupervised}, the stereo matching \cite{martins2018fusion} and the semantic information \cite{zhang2019dispsegnet}. Godard \textit{et al.} \cite{godard2018digging} achieved a significant improvement by compensating for image occlusion. 

Besides estimating depth from the monocular video, \cite{godard2018digging, luo2018every} have proposed to synthesize stereo image pairs for depth estimation. Here the stereo image pairs have been calibrated in advance, the pose network is thus no longer necessary. Depth prediction from this set-up is free of scale ambiguity issue, since the scale information is introduced from the calibrated stereo image pairs.

\noindent {\bf Compensation for Object Motion}\, Most self-supervised monocular depth estimation approaches are subject to rigid scene assumption: scenes captured by video are assumed to be rigid. This assumption is not valid in most autonomous driving scenario, where many moving objects are presented. 

The object motion can be solved by introducing the optical flow map. Yin \textit{et al.} \cite{yin2018geonet} proposed to estimate the residual flow on top of the rigid flow, which is computed from the predicted depth and camera ego-motion. This residual flow can only correct for small error but generally fail for big pixel displacement, \textit{e.g.} when the object is moving fast. Lee \textit{et al.} \cite{lee2019learning} proposed to estimate the residual flow from stereo video. Luo \textit{et al.} \cite{luo2018every} proposed to jointly train networks for depth, camera ego-motion, optical flow and motion segmentation, with enforcing the consistency between each prediction. In \cite{ranjan2019competitive} a similar architecture is adopted, while the system is trained in a competitive collaboration manner. Both \cite{ranjan2019competitive} and \cite{luo2018every} produced State-of-the-art (SoTA) performance of optical flow prediction on KITTI dataset. 

Beyond the scope of self-supervised learning, the estimation of optical flow has been addressed through end-to-end deep regression based methods \cite{dosovitskiy2015flownet, ilg2017flownet}. PWC-Net \cite{sun2018pwc} further improves the efficiency by integrating the pyramid processing and cost volume into their system. Besides optical flow, scene flow \cite{vedula1999three} has been introduced to solve the object motion. Scene flow vector describes the 3D motion of a point. \cite{vogel2013piecewise, yamaguchi2014efficient, menze2015object} estimated the scene flow by fitting a piece-wise rigid representations of motion. They decompose the scene into small rigidly moving plane and solve their motion by enforcing some constraints, like appearance or constant velocity consistency in \cite{vogel2013piecewise}. Battrawy \textit{et al.} \cite{battrawy2019lidar} introduced sparse LiDAR to estimate scene flow together with stereo images. DRISF \cite{ma2019deep} formulates the scene flow estimation as energy minimization in a deep structured model, which can be solved efficiently and outperforms all other approaches. 

In this work we estimate the object motion by modelling it as a rigid-body transform. The scale ambiguity of motion prediction is solved by imposing a geometric constraint loss term. Our network output describes the object movement in 3D space. This is fundamentally different with the work of Casser~\etal~\cite{casser2018depth}, where only an up-to-scale prediction is predicted. This means the magnitude information of motion is missing in their prediction. Neither did they provide the evaluation of the object motion prediction.

\begin{figure*}[!t]
\begin{center}
    \begin{minipage}[t]{.73\textwidth}\vspace*{0pt}
        \includegraphics[width=1.0\linewidth]{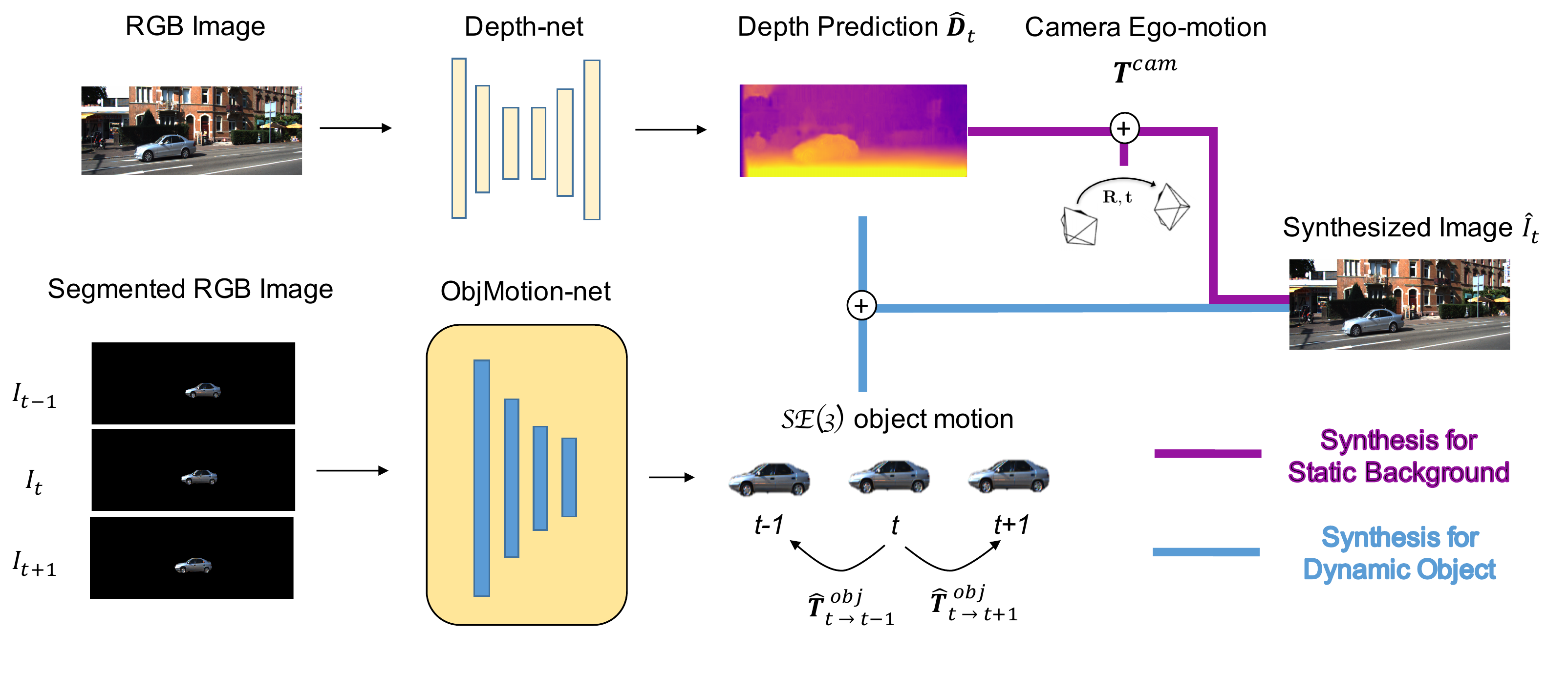}
    \end{minipage}
    \vspace{-4mm}\hfill
    \begin{minipage}[t]{.25\textwidth}
        \vspace{2mm}
        \caption{Framework overview. For view synthesis, each pixel is distinguished as either dynamic or static pixel. Dynamic pixel is synthesized from the individual object motion and depth prediction, while static pixel is reconstructed from the depth and camera ego-motion. The camera ego-motion is pre-computed from the visual odometry library~\cite{Geiger2011IV}. The distinguish of static/dynamic pixel is based on the segmentation mask, provided by Mask R-CNN~\cite{he2017mask}.}
        \label{fig:pipeline}
    \end{minipage}
    \vspace{-2mm}
\end{center}
\end{figure*}

\section{Method}
\label{sec:Method}

We propose a framework for jointly training an object motion network (ObjMotion-net) and a depth network (Depth-net). We firstly explain the view synthesis for dynamic objects, and then provide an overview of our framework. Our networks are supervised by four losses, which are detailed in Sec.~\ref{sec:loss_fun}. 

\subsection{Theory of View Synthesis}
The target frame $I_{tgt}$ is synthesized from the source frame $I_{src}$. For each pixel $p_{tgt}$ in $I_{tgt}$, its correspondence $p_{src}$ in $I_{src}$ is required. The photometric consistency between the synthesized view $\hat{I}_{tgt}$ and its reference $I_{tgt}$ serves as the primary supervision in our system. \\

\noindent {\bf Synthesis for Static Area}\quad Suppose two consecutive frames from a video are given: the target frame $I_{tgt}$ captured at time \textit{t}, and the source frame $I_{src}$ captured at time \textit{t+1}. For pixel $p_{tgt}$ in the static area of $I_{tgt}$, its correspondence $p_{src}$ in $I_{src}$ is computed from Eq.~\ref{eq:view_syn_rigid}:
\begingroup\makeatletter\def\f@size{9}\check@mathfonts
\begin{equation}
\label{eq:view_syn_rigid}
\begin{split}
    h&(p_{src})  \sim K T_{t\rightarrow s}\, X^t(p_{tgt}) \\
    X&^t(p_{tgt})  = \hat{D}(p_{tgt}) K^{-1} h(p_{tgt}) \quad p_{tgt} \in S_0(I_{tgt})
\end{split}
\end{equation}
\endgroup
where $h(p)$ denotes the homogeneous pixel coordinates, $K$ is the camera intrinsics, $T_{t\rightarrow s}$ is the camera ego-motion for the reference system $C_{tgt}$ and $C_{src}$, $X^t(p_{tgt})$ is the projected 3D point of $p_{tgt}$ in the reference system $C_{tgt}$, $\hat{D}(p_{tgt})$ denotes the depth prediction scalar at $p_{tgt}$, $S_0(I_{tgt})$ refers to the static area of $I_{tgt}$. 

\begin{figure}[!t]
\begin{center}
   \includegraphics[width=0.7\linewidth]{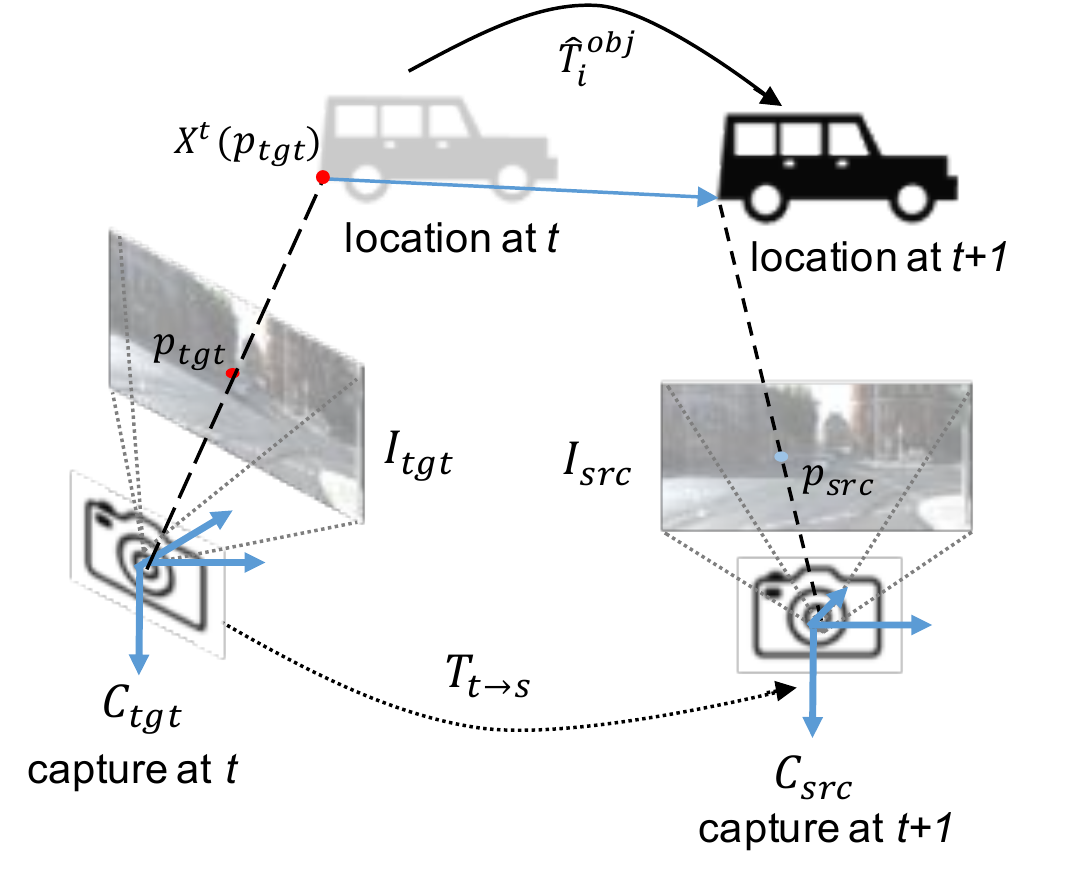}
\end{center}
    \vspace{-2mm}
   \caption{View synthesis for dynamic object. Firstly $p_{tgt}$ is projected into the target camera reference system $C_{tgt}$, denoted as a 3D point $X^t(p_{tgt})$. This point is then transformed by $\hat{T}^{obj}_i$  (for object motion) and $T_{t\rightarrow s}$ (for camera ego-motion), and is finally projected onto $I_{src}$ as the correspondence $p_{src}$. }
    \vspace{-4mm}
\label{fig:dyna_view_syn}
\end{figure}

\noindent {\bf Synthesis for Dynamic Area}\quad Pixel correspondence for dynamic object is computed from Eq.~\ref{eq:view_syn_dyna}. Here the 3D point $X^t(p_{tgt})$ is further transformed by a rigid-body transform $\hat{T}^{obj}_i \in SE(3)$ (6 \textit{dof}, 3 translations and 3 Euler angles). This process is illustrated in Fig.~\ref{fig:dyna_view_syn}.
\begingroup\makeatletter\def\f@size{9}\check@mathfonts
\begin{equation}
\label{eq:view_syn_dyna}
    h(p_{src})  \sim K\, T_{t\rightarrow s}\, \hat{T}^{obj}_i\,  X^t(p_{tgt}) \quad p_{tgt} \in S_i(I_{tgt})
\end{equation}
\endgroup
Here $S_i(I_{tgt})$ refers to pixels in the dynamic area of $I_{tgt}$, whose 3D motion is described as $\hat{T}^{obj}_i$. Suppose there are $n$ moving objects in the scene, we estimate $\hat{T}^{obj}_i\, (i = 1, \dots , n)$ for each individual object. Then the target frame $I_{tgt}$ is synthesized separately for static pixels according to Eq.~\ref{eq:view_syn_rigid}, and for dynamic pixels according to Eq.~\ref{eq:view_syn_dyna}. 

Note we only focus on objects whose movement can be described by a rigid-body transform. These include cars, buses and trucks. Objects like pedestrians are not considered since their movement is too complicated to be described by a 6 \textit{dof} rigid-body transform.

\subsection{Framework Overview}
\label{sec:framework_overview}

Fig.~\ref{fig:pipeline} provides an overview of our framework. It illustrates how the image is synthesized from the network output: depth and object motion for all instances in the scene. We distinguish between the static and dynamic area based on image segmentation mask. The segmentation masks are obtained from the pre-trained Mask R-CNN \cite{he2017mask} model. They highlight instances which move rigidly in the scene.

The segmentation mask also distinguishes between different instances in the scene. We align the instance mask across time, and segment the temporal image sequence by the instance-aligned mask. One masked sequence example is shown as $I_{t-1}$, $I_t$ and $I_{t+1}$ in Fig.~\ref{fig:pipeline}. This serves as the network input to predict the motion $\hat{T}^{obj}_{t\rightarrow t-1}$ and $\hat{T}^{obj}_{t\rightarrow t+1}$ for this specific object. 

In implementation, the actual network prediction is the product of the camera ego-motion and the object motion (\textit{i.e.} $T_{t\rightarrow s} \times \hat{T}^{obj}_{i}$ in Eq.~\ref{eq:view_syn_dyna}). We combine the camera ego-motion and object motion into one single transformation. This transformation is equivalent to a \textit{pesudo} object motion where we assume the camera is static. The \textit{actual} object motion can be decomposed based on the pre-computed camera ego-motion. Combining the camera and object motion together facilitates the employment of geometric constraint loss term (defined in Eq.~\ref{eq:ctrl_signal}), which encodes the magnitude information of object motion.

It is noteworthy that motionless objects are also highlighted by Mask R-CNN. They are treated equally as dynamic objects in our system. It is unnecessary to distinguish between static and dynamic objects, since the input of ObjMotion-net is a segmented image sequence, where only one individual instance is presented. The motion predictions of static objects are equal to the camera ego-motion.

\noindent {\bf Object Motion Network}\, The ObjMotion-net is designed to predict individual object movement. It takes the masked image sequence (shown in Fig.~\ref{fig:pipeline}) as input. All information irrelevant with the target object is excluded. 

The idea of ObjMotion-net is inspired by the Pose-net. Both networks take image sequence as input, and output 6 motion parameters. As shown in \cite{zhou2017unsupervised}, Pose-net is capable to infer the camera ego-motion. This indicates Pose-net can conduct feature extraction and matching which are indispensable procedures for motion inference. We suppose ObjMotion-net, which adopts a similar architecture, also has the capability to extract and match features, and can infer the individual object motion based on these information.

\subsection{Loss Function}
\label{sec:loss_fun}
Our framework employs four loss terms: photometric loss $L_p$, left-right photometric loss $L_{lrp}$, disparity smoothness loss $L_{disp}$ and geometric constraint loss $L_{gc}$.

\noindent {\bf Photometric Loss}\, $L_p$ penalizes the photometric inconsistency between the synthesized view $\hat{I}$ and its reference view $I$. $\hat{I}$ is synthesized based on the prediction from Depth-net and ObjMotion-net, thus $L_p$ provides gradient on both networks. We adopt a robust image similarity measurement SSIM for $L_p$ as formulated in Eq.~\ref{eq: photo_loss}, with $\alpha = 0.85$. The depth is predicted and supervised at multi-scale level to overcome the gradient locality \cite{zhou2017unsupervised}. 
\begingroup\makeatletter\def\f@size{9}\check@mathfonts
\begin{equation}
    L_p = \alpha \frac{1-SSIM(I, \hat{I})}{2} + (1 - \alpha)\|I - \hat{I}\|_1
    \label{eq: photo_loss}
\end{equation}
\endgroup

Note we distinguish the static and dynamic area for the synthesized view $\hat{I}$ when we compute its photometric loss. Instead of averaging the per-pixel photometric difference over the whole image, we average the difference in static and dynamic area separately, and formulate the $L_p$ by summing them. According to~\cite{sfCaoKHM2019}, the separation of photometric loss can compensate the unbalance between the static and dynamic image area, thus provide more supervision signal and contribute to the training of ObjMotion-net. 

\noindent {\bf Left-right Photometric Loss}\, $L_{lrp}$ is imposed to solve the scale ambiguity of the monocular depth prediction. The direct output of our Depth-net is actually the disparity. It can be used to synthesize the left image from its right counterpart, and vice versa. $L_{lrp}$ penalizes the photometric difference of the synthesized stereo images. This provides supervision to solve the scale ambiguity of disparity predictions. 

\noindent {\bf Disparity Smoothness Loss}\, $L_{disp}$ is enforced to penalize a fluctuated disparity prediction. An edge-aware smoothness term is imposed as formualted in Eq.~\ref{equ: l_disp}. Here the disparity smoothness ($\partial_x d$ and $\partial_y d$) is weighted by the exponential image gradient ($e^{ \|-\partial_x I\|}$ and $e^{ \|-\partial_y I\|}$). $x$ and $y$ refers to the gradient along the horizontal or vertical direction. 
\begingroup\makeatletter\def\f@size{9}\check@mathfonts
\begin{equation}
    L_{disp} = |\partial_x d| e^{ \|-\partial_x I\|} + |\partial_y d| e^{ \|-\partial_y I\|}
    \label{equ: l_disp}
\end{equation}
\endgroup

\noindent {\bf Geometric Constraint Loss}\, During experiments we found the translation of object motion tends to be predicted as small values. Similar phenomenon was also observed in \cite{casser2018depth}. We fix this issue by imposing a geometric constraint on the object translation prediction. This constraint provides the magnitude information of the object movement. The geometric constraint $F^{t\rightarrow t+1}_i$ for the $i$-th object between time $t$ to $t+1$ is computed as Eq.~\ref{eq:ctrl_signal}:
\begingroup\makeatletter\def\f@size{9}\check@mathfonts
\begin{equation}
    \begin{split}
        F&^{t\rightarrow t+1}_i = \bar{X}^{t+1}_i - \bar{X}^t_i\\
        \bar{X}&^m_i = |\sum X^m_i(p)| \quad p\in S_i(I_m),m\in \{t, t+1\}\\
        X&^m_i(p) = \hat{D}^m(p) K^{-1} h(p)\quad m\in \{t, t+1\}
    \end{split}
    \label{eq:ctrl_signal}
\end{equation}
\endgroup
$F^{t\rightarrow t+1}_i$ is actually the vector from the 3D object center $\bar{X}^t_i$ to $\bar{X}^{t+1}_i$. Here $|\cdot|$ refers to the mean operator. $X^m_i(p)$ is the projected 3D point of pixel $p$ in the reference system $C_m$, while $S_i(I_m)$ is the $i$-th object area of image $I_m$, with $m\in \{t, t+1\}$ denoting the image capture time.

As mentioned in \ref{sec:framework_overview}, our system predicts a \textit{pesudo} object motion where we assume the camera is static. Ideally the predicted (pseudo) object translation are supposed to be equivalent with the geometric constraint. We impose the geometric constraint loss term $L_{gc}$, which is the L1-norm of the difference between the predicted translation of object motion $\hat{t}_i$ and the geometric constraint $F_i$.
\begingroup\makeatletter\def\f@size{9}\check@mathfonts
\begin{equation}
        L_{gc} = \sum_i^n \| \hat{t}_i - F_i\|_1 \quad \hat{t}_i = [\hat{x}_i, \hat{y}_i, \hat{z}_i]
    \label{eq: ctrl_constraint}
\end{equation}
\endgroup
Here $n$ is the number of instances appeared in the input image sequence. With introducing $L_{gc}$, the issue of the small translation prediction can be fixed. Ablation studies are provided in Sec.~\ref{sec:obj_motion_eval}.

Our final objective is a sum of all loss terms stated above, weighted by their corresponding weight:
\begingroup\makeatletter\def\f@size{9}\check@mathfonts
\begin{equation}
    L_{final} = \lambda_{p} \cdot L_{p} + \lambda_{lrp} \cdot L_{lrp} + \lambda_{disp}\cdot L_{disp} + \lambda_{gc} \cdot L_{gc}
    \label{equ: l_tot}
\end{equation}
\endgroup

\section{Experiments}
\label{sec: experiments}

In this section, we firstly describe the implementation details, and demonstrate evaluation results on individual object motion, disparity, and scene flow prediction. Experiments are conducted on KITTI \cite{geiger2012we}, a dataset provides driving scenes in real-world scenario.

The ObjMotion-net and Depth-net are trained jointly, since the view synthesis for dynamic scene requires both the object motion and depth prediction. However, these two networks can be run independently during test time inference. Their network inputs are irrelevant with each other. 

\subsection{Implementation Details}

\noindent {\bf Dataset and Preprocessing}\, KITTI raw dataset provides videos which cover various scenes. We resize all images into a fixed size $192 \times 640$, and format a temporal image sequence by concatenating $I_{t-1}, I_t$ and $I_{t+1}$ horizontally. 

The evaluation is performed on the training split of KITTI flow 2015 dataset, where the ground truth for disparity, optical flow and scene flow are available. Scenes covered by this training split are excluded during training. 40820 samples and 2070 samples are formatted for training and validation, respectively. Besides the raw dataset, we format another training set from the test split of the multi-view extension of KITTI flow 2015 dataset. Scenes in this split contain more moving vehicles. This contributes to the training of ObjMotion-net. There are 6512 training samples and 364 validation samples in this training set. 

The segmentation mask for image is generated from the pre-trained Mask R-CNN model \cite{he2017mask}. We segment objects which move rigidly in the scene. The instance is aligned according to the Intersection over Union (IoU) of the temporal mask sequence. For example, $M_t^i$ is the mask of instance $i$ at time $t$. Its aligned mask $M_{t-1}^i$ and $M_{t+1}^i$ are obtained by finding the instance mask with the maximum IoU at time $t-1$ and $t+1$. For partially occluded or fast moving objects whose IoU is small, we further check the moving direction of the mask center. We assume the object moving direction between $t-1$ to $t$ and $t$ to $t+1$ (\textit{i.e.}~the 2D vector which connects mask centers) are similar. Aligned masks with significantly different moving direction are discarded. We also ignore very small instances (\textit{i.e.}~the number of mask pixels is less than 400 in one $1392\times 512$ image).

The camera ego-motion is required for view synthesis. Instead of training a pose network, we use the Libviso2 \cite{Geiger2011IV} to estimate the camera ego-motion. 
\\

\begin{figure*}[!t]
\begin{center}
   \includegraphics[width=1.0\linewidth]{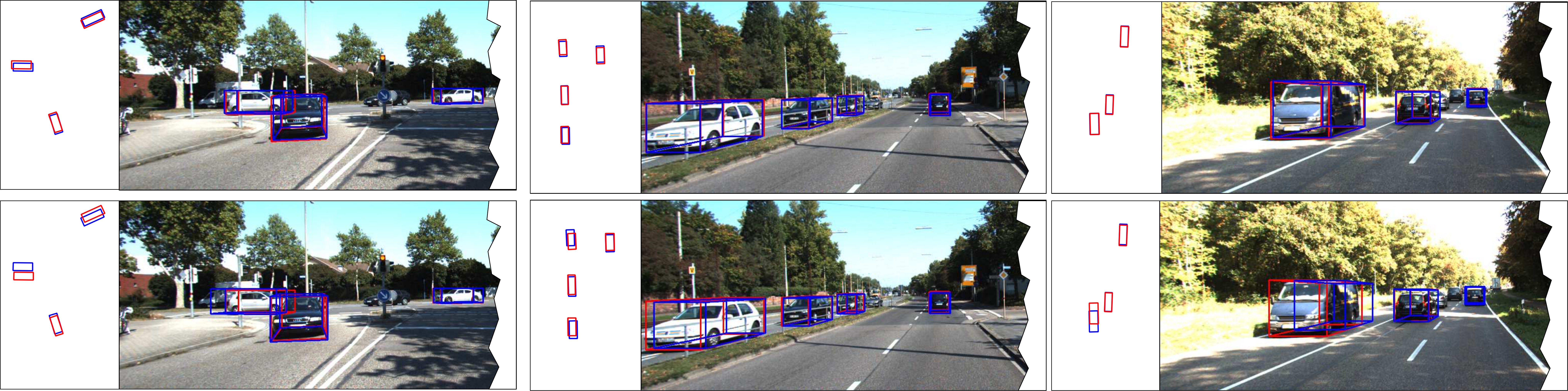}
\end{center}
    \vspace{-2mm}
   \caption{Visualization of Bird's View Box and 3D Bounding Box. Our results (top row) and results from GeoNet \cite{yin2018geonet} (bottom row) are presented. The ground truth is in red while the prediction is in blue. Our predictions have a larger overlapping with the ground truth box. }
%   \vspace{-2mm}
\label{fig:bird_3d}
\end{figure*}

\noindent {\bf Network Architecture}\, Our system contains two sub-networks, the ObjMotion-net and the Depth-net. The ObjMotion-net is designed based on the pose network in \cite{yin2018geonet}. We adopt ReLU~\cite{nair2010rectified} activation for all convolutional layers. Batch normalization (BN)~\cite{ioffe2015batch} is excluded in ObjMotion-net, since experiments demonstrate BN does not contribute to the performance. 

For the Depth-net, we adopt the architecture in \cite{yin2018geonet} as backbone. This structure consists of the encoder and the decoder part. The basic structure of ResNet50 is adopted for the encoder. While in decoder the combination of convolution and upsampling is used for upscaling the feature map. Skip connections between the encoder and the decoder are added to integrate global and local information. ReLU and BN are adopted for all layers of Depth-net except for the prediction layer, where the Sigmoid activation is used and BN is excluded. 

\subsection{Training Details}
Our system is implemented using TensorFlow framework \cite{abadi2016tensorflow}. Color augmentation is performed on the fly. The network is optimized using Adam optimizer, with $\beta_1 = 0.9$ and $\beta_2 = 0.999$ respectively. Our system is trained on a single TitanXP GPU. A stage-wise training strategy is adopted, with training Depth-net alone at the beginning, and then jointly training Depth-net and ObjMotion-net.

\noindent {\bf Training Depth-net}\, We firstly train the Depth-net, since accurate depth prediction is necessary to compute the geometric constraint for the ObjMotion-net. According to Godard~\textit{et al.}~\cite{godard2017unsupervised}, pre-training network on Cityscapes dataset~\cite{Cordts2016Cityscapes} contributes to the performance of Depth-net. We trained the Depth-net on the Cityscapes dataset for 600K steps. The training was supervised by the left-right photometric consistency $L_{lrp}$ ($\lambda_{lrp} = 1.0$) and disparity smoothness $L_{disp}$ ($\lambda_{disp} = 0.5$). The learning rate and the batch size are 0.0001 and 2, respectively. The photometric loss $L_p$ is excluded in this stage, since we do not model object motion. Employment of $L_p$ will lead to inaccurate depth prediction in dynamic area. This issue is also observed by Luo~\etal~\cite{luo2018every}.

After the pre-training on Cityscapes, we continued to train the Depth-net on the formatted KITTI dataset for 500K steps. All hyper-parameters were kept same except for the $\lambda_{disp}$, which we changed to 25.0. We found in experiments that a higher smoothness penalization was indispensable, otherwise the disparity prediction became unreasonably fluctuated.

\noindent {\bf Training ObjMotion-net with Depth-net}\, We then jointly optimize the Depth-net and ObjMotion-net. Besides the $L_{lrp}$ and the $L_{disp}$, the photometric loss $L_{p}$ and geometric constraint loss $L_{gc}$ are imposed. The loss weights are set to be $\lambda_{p} = \lambda_{lrp} = 1.0$, $\lambda_{disp} = 25.0$, and $\lambda_{gc} = 1.0$. The learning rates for Depth-net and ObjMotion-net are 0.0001 and 0.0002, respectively. And the batch size is 2. 

After training on the KITTI raw dataset for 200K iteration, we fine-tune the ObjMotion-net on the test split of KITTI Flow dataset 2015, with fixing the parameters of Depth-net. All hyper-parameters remain the same. The ObjMotion-net is trained for 100K iterations in this stage. 

\begin{table}[t]
\vspace{-2mm}
\begin{center}
\resizebox{0.31\textwidth}{!}{
\begin{tabular}{|l|cc|}
\hline
Method & Bird View & 3D Box  \\
\hline
    CC~\cite{ranjan2019competitive} & 43.10\% & 43.60\%  \\
    GeoNet \cite{yin2018geonet} & 57.54\% &  56.00\%  \\
    Ours (no $L_{gc}$) & 45.02\% & 43.67 \% \\
    Ours & \textbf{72.31\%} & \textbf{70.61\%}  \\
\hline
\end{tabular}}
\end{center}
\caption{Average IoU of bird's view box and 3D bounding box.}
\label{tab: IoU}
\vspace{-4mm}
\end{table}

\subsection{Individual Object Motion Evaluation}
\label{sec:obj_motion_eval}
Our system predicts individual object motion in 3D space. To demonstrate the effectiveness of our system, we present the IoU of the bird's view box and 3D bounding box. Take the example of 3D bounding box: the 3D bounding box for object $i$ at time $t$, denoted as $B^i_{t}$, is transformed by its object motion prediction $\hat{T}^i_{t+1}$. Then the predicted location of box at time $t+1$, $\hat{B}^i_{t+1}$ is obtained. We compute the IoU between $\hat{B}^i_{t+1}$ and its ground truth $B^i_{t+1}$. The average IoU indicates the performance of our ObjMotion-net. 

We evaluate on 80 temporal image pairs which are contained in both the training split of KITTI tracking (provides ground truth bounding box) and flow 2015 dataset. Objects are segmented by the pre-trained Mask R-CNN model \cite{he2017mask}. We do not use the ground truth segmentation in flow 2015, since it only provides the segmentation for $I_t$, while the segmentation for $I_{t-1}$ and $I_{t+1}$ are necessary for motion prediction. In total 204 objects are selected from these 80 image pairs. 

We compare our results with GeoNet~\cite{yin2018geonet} and CC~\cite{ranjan2019competitive}. Both approaches predict dense optical flow and depth map, from which we can compute the pixel-wise scene flow vector in 3D space. The individual object motion can then be inferred by averaging over the object segmentation mask on the scene flow map. Their up-to-scale depth prediction are scaled by the median ground truth depth. 

In Fig.~\ref{fig:bird_3d}, we compare our qualitative results with GeoNet~\cite{yin2018geonet}. It can be seen that our predicted bounding boxes have a higher overlapping with the ground truth. Quantitative results in Table~\ref{tab: IoU} also show our system has a higher average IoU, with 72.31 \% for bird's view box and 70.61\% for 3D bounding box. This demonstrates the effectiveness of our system to predict the individual object motion. We report the evaluation result without geometric constraint loss $L_{gc}$. It can be seen that without $L_{gc}$ the performance is significantly worse. The lower IoU of CC~\cite{ranjan2019competitive} prediction results from its inaccurate depth prediction.

Besides IoU of bounding box, Cao~\textit{et al.}~\cite{sfCaoKHM2019} proposed to evaluate the motion speed and direction. However, they did not publish their models and evaluation details. Thus it is not possible to compare with them. 

\begin{figure*}
\begin{center}
\begin{minipage}[t]{.75\textwidth}\vspace*{0pt}
\includegraphics[width=.98\linewidth, height=6.6cm]{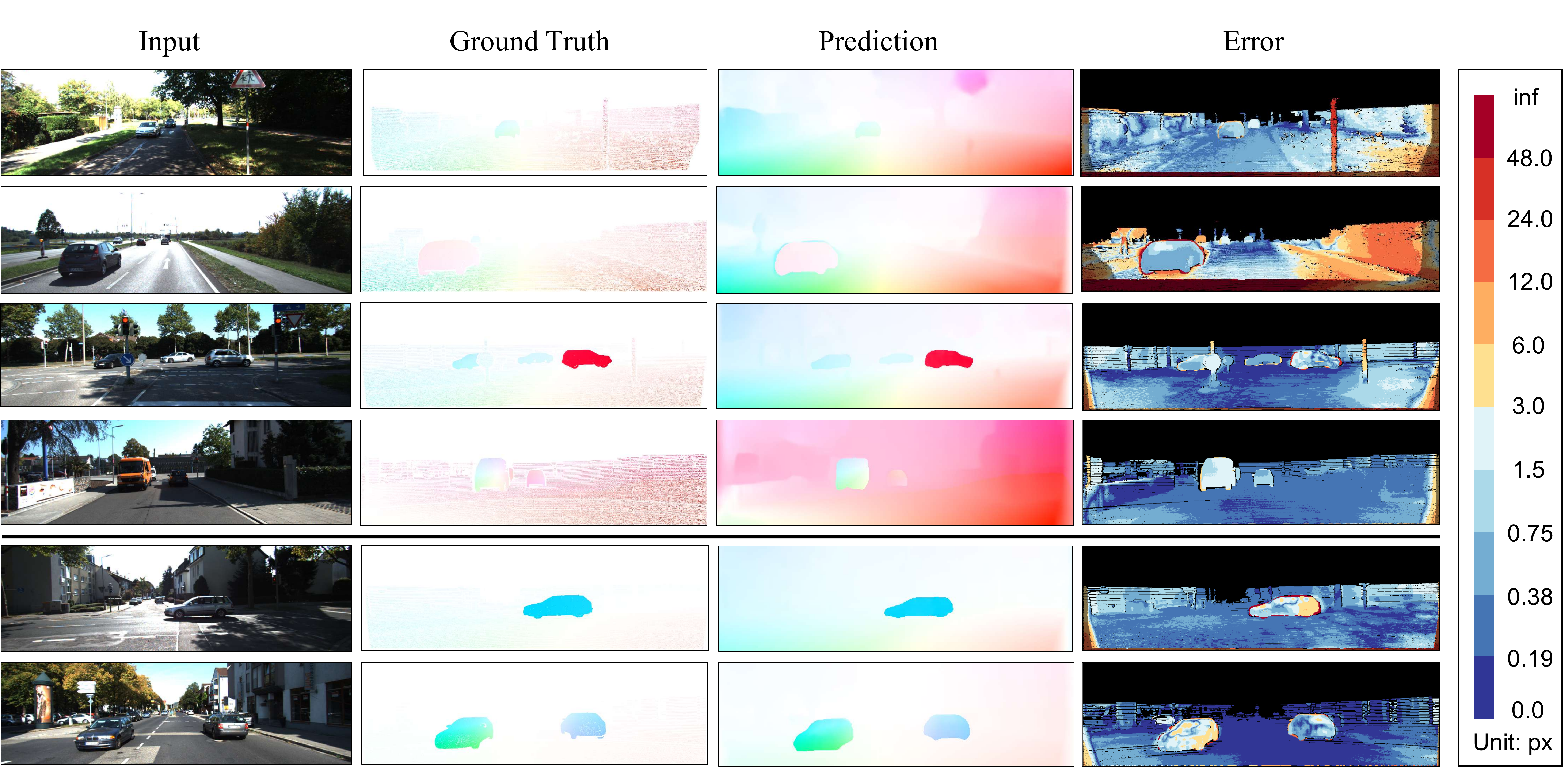}
\end{minipage}
\vspace{-3mm}\hfill
\begin{minipage}[t]{.23\textwidth}
\vspace{4mm}
\caption{Visualization of optical flow prediction for KITTI flow 2015 training split. The error magnitude is encoded into different colors according to the legend at the right-hand side. Basically good pixel is in blue while bad pixel is in orange/red. The top four rows show some successed samples, where the error of most pixels in dynamic region are below 3px. The bottom two rows show two imperfect examples. The bias of bad pixel is slightly over the threshold due to the imperfection of view synthesis.}
\label{fig:flow_img}
\end{minipage}
\vspace{-3mm}
\end{center}
\end{figure*}

\begin{table}[t]
\begin{center}
\resizebox{0.4\textwidth}{!}{
\begin{tabular}{|l|ccc|}
\hline
\multirow{2}{*}{Method} & \multicolumn{3}{c|}{Bad Pixel Percentage} \\\cline{2-4}
                        & bg & fg & all \\
\hline
    CC~\cite{ranjan2019competitive} & 35.03\% & 42.74\% & 36.20\% \\
    Monodepth2 \cite{godard2018digging} & \textbf{18.60\%} & 44.47 \% & \textbf{22.50}\%    \\
    EPC++ \cite{luo2018every} & 22.76\% & 26.63\% & 23.84 \% \\
    Ours (1st stage) & 29.49\% & 19.62\% & 28.00\%  \\
    Ours & 28.08\% & \textbf{16.65\%} & 26.36\% \\
\hline
\end{tabular}}
\end{center}
\caption{Bad pixel percentage of disparity prediction}
\label{tab: disp}
\vspace{-3mm}
\end{table}

\subsection{Monocular Disparity Evaluation}
To demonstrate the contribution of ObjMotion-net towards disparity estimation (in particular for dynamic area), we evaluate on the training split of flow 2015 dataset and report the average bad pixel percentage (BPP) of disparity prediction. A pixel is considered as bad if its prediction error $\geq$ 3px or $\geq$ 5\%. Besides, the ground truth segmentation masks for moving objects are provided. This makes it possible to evaluate within dynamic region (\textit{fg} in Table~\ref{tab: disp}), which in our case is the Region-of-Interest. The BPP for static background (bg) and overall area (all) are also presented.

Our result achieves SoTA performance in terms of BPP in fg. We produce the lowest value of $16.65 \%$ in foreground, which is better than BPP of Monodepth2 ($44.47\%$) and EPC++ ($26.63\%$). It is important to note that the published models of other works (CC, Monodepth2 and EPC++) have been trained on the test images, since they adopted another training split (Eigen split~\cite{eigen2014depth}). Our system did not witness the test images, but still produce a better performance in foreground and comparable result in overall area. This demonstrates the disparity prediction for dynamic objects can be improved through modelling the object motion explicitly.

\subsection{Scene Flow Evaluation}

We present the evaluation results on 3D scene flow in Table~\ref{tab: sceneflow}. The evaluation conducts on the predicted disparity for two consecutive frames: $\hat{D}(I_t)$ and $\hat{D}(I_{t+1})$, and the 2D optical flow map $\hat{F}_{t\rightarrow t+1}$. In our system, we do not have a component to explicitly predict the pixel-wise optical flow. The optical flow prediction is obtained through view synthesis with taking the object motion into account. We present the visualization of our optical flow results in Fig.~\ref{fig:flow_img}.

\begin{table}[t]
\begin{center}
\resizebox{0.42\textwidth}{!}{
\begin{tabular}{|l|ccc|}
\hline
Method & bg & fg & all \\
\hline
GeoNet \cite{yin2018geonet} & 66.8\% & 90.4\% & 70.7\% \\
Mono + Geo & 39.4\% & 70.9\% & 44.7\% \\
EPC++ \cite{luo2018every}  & $>$ 22.8\% & $>$ 70.4\% & $>$ 60.3\% \\
CC~\cite{ranjan2019competitive} & 50.2\% & \textbf{60.0\%} & 51.8\% \\
Ours  & 38.2\% & 65.9\% & \textbf{42.8\%} \\
\hline
\end{tabular}}
\end{center}
\caption{Bad pixel percentage of scene flow prediction. For the evaluation of \textit{Mono + Geo}, the disparity from Monodepth2 \cite{godard2018digging} and the flow prediction from GeoNet \cite{yin2018geonet} are used.} 
\label{tab: sceneflow}
\vspace{-5mm}
\end{table}

Our result achieves the best BPP in overall area (42.8\%), compared with 44.7\% from Mono + Geo, and 51.8\% from CC. In foreground, our BPP (65.9\%) is worse than the results of CC (60.0\%). This is because our flow results are synthesized from depth and object motion prediction. Any subtle bias in view synthesis (like camera intrinsics, depth, object motion) may result in an error larger than the bad pixel threshold (3px). We present two imperfect flow prediction examples in the last two rows of Fig.~\ref{fig:flow_img}. Although some bad pixels are presented in the dynamic area, the magnitude of their bias are barely over the threshold due to the imperfect view synthesis. 

Nevertheless, our system is capable to capture the holistic object motion. Our system produces the lowest BPP in overall area, and it can be seen from Fig.~\ref{fig:flow_img} that the bias of most dynamic pixels are below the bad pixel threshold. 

\section{Conclusion}

We have presented a self-supervised learning framework for individual object motion and depth estimation. The object motion is modelled as a rigid-body transformation. Our system is able to learn the object motion from unlabelled video. This contributes to scene flow prediction, and improve the depth estimation in dynamic area of the scene. 

It would be interesting to explore the following questions in future: 1) Integrate scale information from other sources. In our system the scale information of object motion is extracted from the depth prediction (the \textit{control signal}). It would be difficult to apply our system in case where the depth prediction is not reliable. In future we can try to integrate scale information from other sources, like stereo image pairs, or sparse depth \textit{ground truth} from Lidar. 2) Estimate the motion of pedestrians. Currently we focus on the object motion which can be described by a rigid-body transformation. While non-rigid motion, like the movement of pedestrians, is also common in driving scenario. We may dissect pedestrians into smaller parts which is moving rigidly, or try to model its motion in a different way. 

{\small
\bibliographystyle{ieee_fullname}
\bibliography{egbib}
}

\end{document}